\RequirePackage{fix-cm}
\documentclass[twocolumn]{svjour3}          
\smartqed  
\usepackage{graphicx}
\usepackage[numbers]{natbib}
\usepackage{subfig}
\usepackage{amsmath}
\usepackage{color}
\usepackage{url}
\usepackage{mathrsfs}
\usepackage{lmodern}
\usepackage{verbatim}
\usepackage{mathtools}
\usepackage{amsmath}
\usepackage{algpseudocode}

\usepackage[ruled,vlined]{algorithm2e}
\begin{document}

\title{Optical Braille Recognition using Circular Hough Transform}


\author{
	Zeba Khanam$^*$  \and
	Atiya Usmani    
}


\institute{	$^*$Corresponding Author \\ 
	Zeba Khanam \at
	School of Computer Science and Electronic Engineering, \\
	University of Essex,\\Essex,U.K.\\ 
	\email{zeba.khanam@essex.ac.uk} 
	\and
	Atiya Usmani \at
	Insight Centre for Data Analytics,\\
	NUI, Galway, Ireland.\\
	\email{atiya.usmani@insight-centre.org}  }

\date{Received: date / Accepted: date}

\maketitle

\begin{abstract}
Braille has empowered visually challenged community to read and write. But at the same time, it has created a gap due to widespread inability of non-Braille users to understand Braille scripts. This gap has fuelled researchers to propose Optical Braille Recognition techniques to convert Braille documents to natural language. The main motivation of this work is to cement the communication gap at academic institutions by translating personal documents of blind students. This has been accomplished by proposing an economical and effective technique which digitizes Braille documents using a smartphone camera. For any given Braille image, a dot detection mechanism based on Hough transform is proposed which is invariant to skewness, noise and other deterrents. The detected dots are then clustered into Braille cells using distance-based clustering algorithm. In succession, the standard physical parameters of each Braille cells are estimated for feature extraction and classification as natural language characters. The comprehensive evaluation of this technique on the proposed dataset of 54 Braille scripts has yielded into accuracy of 98.71\%.

\end{abstract}

\section{Introduction}
\label{intro}
Written communication has played a pivotal role in development of human civilization. It has emerged as a medium for humans to communicate and share ideas and knowledge beyond the bound of time and space. This power of communication was only possessed by people of sight until recently. In 1829, Braille, a tactile writing system, was proposed which empowered visually impaired population to read and write. Since its inception, Braille has been extensively used by the blind community for all forms of written communication. 

\begin{figure}[t]
	\centering
	
	\includegraphics[height=0.3\textheight]{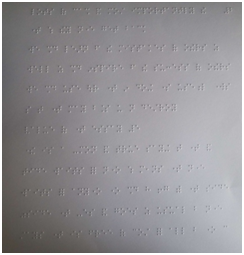}
	\caption{A sample of single-sided Braille document}
	\label{fig:3}       
\end{figure}

A Braille script is a structured document that records all the information in form of protrusions and depressions, a sample is illustrated in Figure \ref{fig:3}. Each character, referred as a Braille cell, is a cluster of six dots arranged as an array of three rows and two columns. Each cell is standardized i.e. radius and height of the dot and cell respectively, inter-cell and intra-cell dot distances are fixed, as shown in Figure \ref{fig:17}.  Each dot can either be embossed as a protrusion (raised) or be flat leading to $2^6 = 64$ combinations. If all the dots are flat, it signifies the area is a blank space. Therefore, a Braille cell can represent 63 characters and symbols. The encoding of Braille cells to the natural language depends on the grade of Braille. Grade 1 Braille has one-to-one mapping between both the representations. This implies that a single Braille cell corresponds to an individual character of the natural language. On the other end of spectrum, Grade 2 Braille has one-to-many mapping. Here, a single Braille character can correspond to a string of natural language characters.

The combined thickness of page material and protrusions make Braille documents bulky. Considerable amount of personal documents are single-sided (printed on one side) however one can find elaborate piece of literature and dictionary printed as double-sided. The double-sided Braille documents are able to concise the text by recording protrusions and depression on both the sides. 

There have been substantial efforts to maintain uniformity across the languages (English, French, Hindi etc.) and the domains (Language, Mathematics, Music etc.). But there prevails incompatibility for instance  variation in cardinality of set of alphabets across the spectrum. Therefore, different convention is followed for each language and domain. 

Beyond any doubt, Braille has emancipated visually impaired population to step out from the dark world of illiteracy to illumination of knowledge but, another side of the coin is that it has dug a trench around the world of Braille writings. This gap stems from non-readability of Braille scripts by the non-Braille users and has crippled the visually impaired people in their communication. The main victims of this written communication hindrance are blind students. They are always in dire need of  writing scribes to submit assignments and take exams. This inhibits severe effect on their academic performance and independence. We strive to bridge this existing communication gap in this paper by proposing a robust approach called as Optical Braille Recognition (OBR).  This approach translates Braille academic documents for non-Braille users. 

\begin{figure}[h]
	\centering
	
	\includegraphics[scale=0.3]{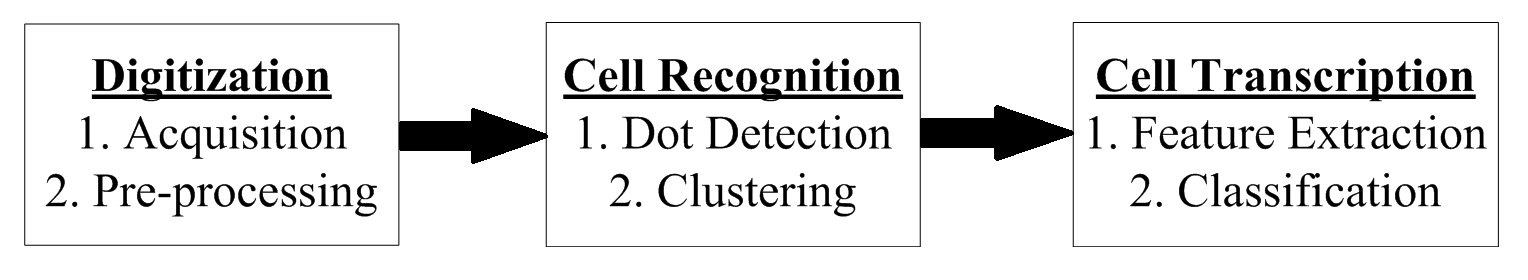}
	\caption{Flowchart illustrating the OBR algorithm}
	\label{fig:4}       
\end{figure}

\begin{figure}[t]
	\centering
	\subfloat[]{
		\includegraphics[height= 0.2\textheight]{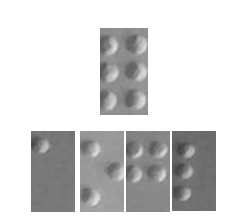}} \\
	\subfloat[]{
		\includegraphics[height=0.3\textheight]{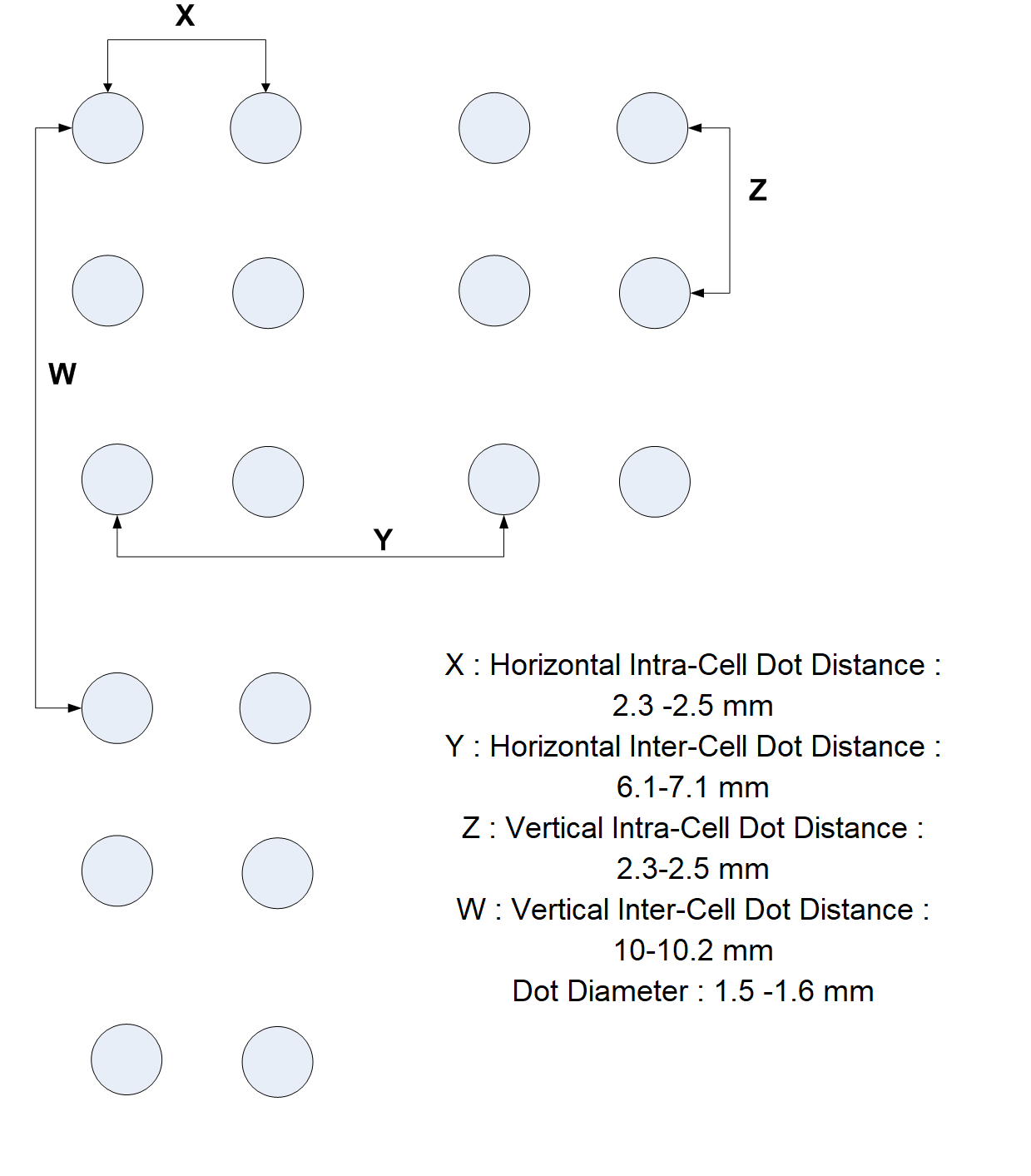} } 
	\caption{(a) Sample Braille cells  (b) Standard Braille cell measurements }
	\label{fig:17}
\end{figure}

The proposed OBR algorithm is a \textit{three-step} process: Digitization, Cell Recognition and Cell Transcription, as shown in Figure \ref{fig:4} 
The \textit{first step} attempts to digitize Braille documents using cameras and scanners. The absence of any visually contrasting information in the documents are major hindrance in acquisition of  the image of Braille documents. This had prompted initial works to deploy complex setup of cameras and light sources \citep{dubus1988image}, \citep{franccois1985reproduction}, \citep{hentzschel1993optical} for homogeneous illumination. \citep{kitchings1995analysis}, \citep{mennens1994optical} are few early works to use scanners. Due to ease of availability and economical reasons, scanners are used predominately. 

This is followed by \textit{second step}, Cell Recognition, which segments a captured Braille document into individual Braille cells. The previous research efforts in this direction have exploited use of basic image processing techniques like correlation with a mask \citep{mennens1994optical}, horizontal and vertical histogram peaks \citep{wajid2011imprinted},\citep{alsalman2012novel}. Few works have localized braille dots using lines \citep{padmavathi2013conversion} and grids \citep{antonacopoulos2004robust},\citep{falcon2005image} based on X and Y projections. For the case of double sided Braille documents, three-level grey threshold has been used to identify protrusion, depression and background in the image \citep{antonacopoulos2004robust},\citep{falcon2005image}.

In the end \textit{third step}, Cell Transcription converts Braille cells into their equivalent natural language characters. Recent advances in machine learning have also been extensively used to achieve cell transcription. Prominent works like Mamba et al. \citep{namba2006cellular} have used a Cellular Neural Network (CNN) to recognize Braille characters and Ting Li et. al \citep{li2014deep} has used greedy layer-wise pre-training algorithm \citep{hinton2006fast} for feature extractor and stacked denoising Auto Encoder (SDAE) \citep{vincent2010stacked} to clean the partially corrupted input and extract the anti-noise feature. Z. Tai et. al \citep{tai2010braille} has used Belief Propagation assuming the document to be a Hidden Markov Model. This work is worth noting as it successfully estimated  highly adaptive parameters of Braille documents. Therefore, it emerges as a possible candidate for comparison with our work.
 
All the above mentioned systems and techniques are non-portable and their practical relevance is questionable. A handful works have tried to proceed in this direction using a mobile phone. Zhang et al. \cite{zhang2007braille} digitizes Braille documents using an embedded camera  of mobile phone. The proposed technique took advantage of standard structure of  Braille documents by comparing distance between the dots to locate dots. The major loophole of this technique is that inter-cell and intra-cell dot distances are subjected to change due to skewness, noise and rotational changes which may creep in during image capture.  

In this paper, we have proposed a portable Braille recognition which toes a different line as compared to the existing works. The prospective stakeholders of our proposed approach are primary school blind students thus we have selected English Grade 1 single-sided Braille documents. Unlike previous works, we have digitized the Braille documents using an eight megapixel smartphone camera. On the acquired Braille images, we apply a series of pre-processing algorithms to de-noise the image, remove skewness and enhance the dots. The Braille dots are then detected using Hough transform. The detected dots are clustered using a novel distance based clustering algorithm into Braille cells. This is followed by extraction of features from Braille cells to translate the documents. The contributions of this paper can be summarized as:
\begin{enumerate}
	\item Development of a comprehensive dataset comprising 54 English Grade 1 single-sided Braille documents. To the best of our knowledge, as no such standard dataset is available in public domain.
	\item  Hough transform based Braille dots detection mechanism which exploits circular shape of dots. 
	\item A novel distance based Braille cell clustering technique and robust feature extraction methodology.
	\item The experimental results demonstrates efficacy of the proposed approach. Particularly, the overall accuracy obtained is 98.71 \% which outperforms the state-of-the-art techniques.  
\end{enumerate}

 The rest of paper has been organized as follows. The next section elaborates the proposed methodology in detail. This is followed by comprehensive analysis of the proposed methodology based on the experimental observations and results in Section 3. In Section 4 to tie the loose knots, the paper is concluded by summarizing research and also shedding  light on the future prospects.

 \section{Proposed Method}
 \label{methods}

 In perpetuation with the above discussion, proposed OBR algorithm can be described as a three step process (refer Figure \ref{fig:4}). The first step,\textit{ `Digitization'} consists of two main sub steps. At first, the Braille document is converted into a raw image using a camera and then a series of pre-processing techniques are applied to enhance the dots.\textit{ `Cell Recognition'}, the second step mainly focuses on segmentation and clustering of Braille dots from the document. The final step i.e. \textit{ `Cell Translation'}, extracts features and classifies Braille cells to readable natural language characters. We will now discuss  each of the steps in details in the following sub-sections.

 \subsection{Digitization }
 \label{step1}
 
 Recent time has witnessed an upsurge in cameras and smartphones. Digitization of texts using smartphone and camera is recurrent. The main objective of our proposed approach is to enhance portability. Thus, we have taken advantage of integration of technology with our society. The initial requirement of any OBR technique is image acquisition of Braille documents which is achieved using an eight megapixel camera of a mobile phone.

 When the image is acquired using the camera, the document is illuminated from an offset angle. This allows many undesired artifacts like false shadow, noise and skewness to fuse in the acquired image. Many of these artifacts results from dearth of any visual and colour contrasting information in the document. 
 
  Many approaches in past have tried to remove these deters using pre-processing techniques like median filtering, dilation, contrast stretching as pre-processing step \citep{isayed2015review} \citep{wajid2011imprinted}. We amalgamate series of pre-processing techniques which not only focuses on removal undesired noise but also enhances dots to improve efficiency of the forthcoming steps of our proposed OBR approach. The employed pre-processing techniques are :
 
 \begin{enumerate}
 
 	\item \textbf{Grayscale Thresholding} : Due to the nature of Braille documents and inhomogeneous illumination,  the colour depth of raw images is converted to 8-bit (greyscale) from 24-bit (RGB). This allows us to get rid of various deficiency pertaining to low quality of paper, annotation and stamps.   
 	\item \textbf{Median Filtering}: As the outcome of grey thresholding, Braille document appears with mid-gray background. This is due to a set of noises introduced while the image is captured which can be modelled as \textit{salt and pepper} noise. Median Filtering is perfect candidate for such noise removal. It effectively removes salt and pepper noise along with other kinds of noises, if any.
 	\item \textbf{Binarization}: After removing the noise, the grayscale image is converted into a binary image to further differentiate Braille dots from the background.
 	\item \textbf{Complement}: It is observed at this stage that colour of Braille dots (black) and the background (white) are complimentary. Therefore, to enhance the performance of our Braille dot segmentation, we highlight white Braille dots on black background.     
 	\item \textbf{Dilation}: The shape of Braille dots is further enhanced using a morphological image processing technique i.e. `Dilation'.
 \end{enumerate}
 
\begin{figure}
 	\centering
 	
 	\includegraphics[height=0.3\textheight]{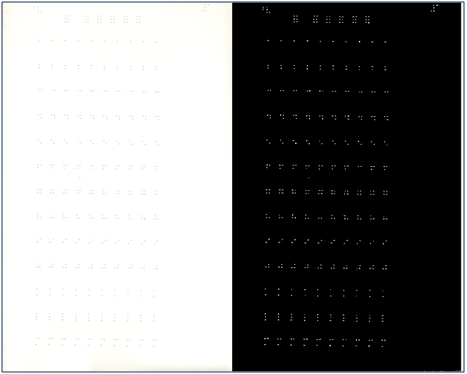}
 	\caption{The left image shows the raw input image and the right image is obtained after pre-processing phase}
 	\label{fig:5}       
 \end{figure}

 Application of all the pre-processing steps on the raw acquired image has been represented in Figure \ref{fig:5}.
 
\subsection{Cell Recognition}
\label{step2}
As the second step of our approach, \textit{`Cell Recognition'} localizes  Braille dots and clusters them into a Braille cell. At first, we will focus on localization of Braille dots. 

The outcome of all the pre-processing techniques (Section 2.1) on the acquired image allows Braille dots to manifest itself as white region on black background (refer Figure \ref{fig:5}). For any given Braille document, dots are embossed as circular protrusions. In past, Hough transform has performed effectively for circle detection \citep{illingworth1987adaptive}, \citep{illingworth1988survey} and \citep{yuen1990comparative} in image processing. Hence, this technique has been judiciously adopted for Braille dot localization due to their standard circular shape. 

\begin{figure}
	\centering
	
		\includegraphics[height= 0.275\textheight , width=0.75\linewidth]{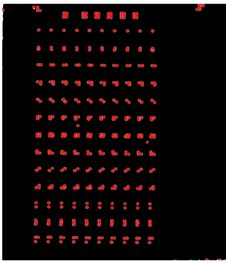} \\
	\caption{Results of Braille dot detection }
	\label{fig:6}
\end{figure}

Let us assume that $I$ is the acquired image and $\mathscr{P}( )$ is a transform which applies pre-processing techniques on the input image. Hence, $I'$ processed image can be obtained using : 
\begin{equation}
\qquad \qquad \qquad \qquad I' = \mathscr{P}(I)
\end{equation}
Let us also assume $\mathscr{H}( )$ is a function which applies Hough transform on the given input image ($I'$) and returns two matrices denoting the coordinates of centre ($D_c$) and radius ($D_r$) of detected Braille dots. Therefore, this transform can be denoted as :
	\begin{equation}
	 \qquad \qquad \qquad \qquad D_c ,  D_r = \mathscr{H}(I')
	 \label{e1}
	\end{equation}
The detected Braille dots using Hough transform can be visualized by plotting detected circles on the pre-processed image ($I'$) as illustrated in Figure \ref{fig:6}.

 After localization of Braille Dots, the next step is to cluster them into Braille cells. By the set norms of the Braille society, Braille dots and cells should always exhibit prominent physical characteristics. While proposing a clustering strategy, a meticulous evaluation allowed us to impose a clustering algorithm which will be based on \textit{intra-cell and inter-cell dot distances} (refer Figure \ref{fig:17}). A single page contains multiple Braille cells but there is an inconsistency in intra-cell and inter-cell dot distance due to skewness and other undesired artifices (discussed in Section 2.1). Thus, we aim to find out the maximum intra-cell and inter-cell dot distances for any Braille page. We define two variables: $HOR_{max}$ as maximum horizontal intra-cell distance and $VER_{max}$ as maximum vertical intra-cell distance, respectively. 
 \subsubsection{Calculation of $HOR_{max}$ and $VER_{max}$:} 
In order to calculate these two variables, we will define two matrices $HOR_{near}$  and $VER_{near}$ which contain horizontal and vertical distance respectively, for all dots with its nearest neighbour. The intra-cell dot distance is distance between any two dots belonging to a single Braille cell. If we calculate distance between any two dots, it is expected to be minimum if the dot belongs to same Braille cell (imposed physical characteristic). 
 For any dot say $dot_i$, its nearest neighbour is $dot_j$. \textit{Iff}, distance between $dot_i$ and $dot_j$ is \textit{minimum} as compared to other existing dots. The above two matrices are mathematically represented as follows :
\begin{equation}
\label{eq:1}
\qquad  HOR_{near} = [H_D (dot_i,dot_j) ; \forall \text{ }dot_i \in D_c]
\end{equation}
\begin{equation}
\label{eq:2}
\qquad VER_{near} = [V_D (dot_i,dot_j) ; \forall \text{ }dot_i \in D_c]
\end{equation}
where $H_D(dot_i,dot_j)$ and $V_D(dot_i,dot_j)$ are the distance operators which calculate horizontal and vertical \textit{city block distance} between $dot_i$ and $dot_j$, respectively and can be formulated as follows:
 \begin{equation}
 \qquad \qquad  H_D(dot_i,dot_j) = | {dot_i}^x - {dot_j}^x | 
 \label{e3}
 \end{equation}
  \begin{equation}
 \qquad \qquad V_D(dot_i,dot_j) = | {dot_i}^y - {dot_j}^y | 
 \label{e4}
 \end{equation}
It is worth considering that nearest neighbour dot of any dot will generally belong to the same Braille cell. But in certain cases like letter `A' which has only one dot, nearest neighbour dot will belong to the adjacent Braille cell.
If we will analyse the histograms of $H_{near}$ and $V_{near}$, we will find two or more peaks. The first peak of histogram signifies varying intra-cell dot distance across the given Braille page. $HOR_{max}$ and $VER_{max}$ will be the maximum distance in the first peak of the histograms. 

\begin{figure}
	\centering
	
	\includegraphics[scale=0.75,width =0.5\textwidth]{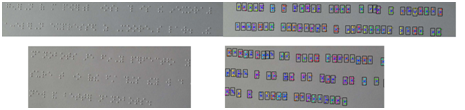}
	\caption{Clustered Braille cells in skewed documents}
	\label{fig:20}       
\end{figure}

After obtaining $HOR_{max}$ and $VER_{max}$, we have found the thresholds for our clustering algorithm.
The proposed Braille cell clustering approach has been described algorithmically using Algorithm \ref{euclid}. 
Algorithm \ref{euclid}\textit{ clusters} the detected dots $D_c$ obtained using equation \ref{e1} into \textit{a set of Braille cells $C$}. Each Braille cell $c_i \in C$ is a set of dots $d_z \in D_c$ and can be formulated as:
\begin{equation}
\label{e6}
\qquad \qquad \qquad \qquad c_i = \{ d_1\dots d_z ; 1 \le z \le 6 \}   
\end{equation}
 \begin{algorithm} [!h]
	
	\small{
		\KwIn{ \\ 1. $D_c$ : Centre Cordinates of dots detected \\ 
			2. $HOR _{max}$ , $VER_{max}$ : Horizontal and Vertical maximum distance, respectively 
		}
		\KwOut{$C = \{c_1,c_2,\dots,c_w\}$ : Set of Braille cells }
		\Begin{
			\textbf{Initialization:} \\
			\begin{enumerate}
				\item $C = NULL; $
				 \item  $K  = |C|$
					\Comment{K denotes number of Braille cells in C} 
			
			\end{enumerate}
			\For{ each $dot_{i} \in D_c$}
			{
			\For{each $dot_{j}~\forall j \in \{i +1 \dots |D_c| \}$}
			{ 
                 Calculating inter-dot distances using Equations \ref{e3} and \ref{e4};  \\
                \textit{ \{$\hat{h}$ and $\hat{v}$ represents horizontal and vertical distances between $dot_i$ and $dot_j$, respectively\}} \\
				\If{$\hat{h} \le HOR_{max}$ \textbf{AND} $ \hat{v} \le VER_{max}$ }{
					\eIf{($C == NULL $) \textbf{OR} ( $dot_i$ \textbf{AND}  $dot_j \notin C $)}{
						$c_{K+1}$ $ <= $ $\{dot_i, dot_j\}$; \\
						Creating a new Braille cell $c_{K+1}$ containing $dot_i$ and $dot_j$; \\
					}{
				    \eIf{$dot_{i} \in C$ \textbf{OR} $dot_{j} \in C$}{
				     Find Braille cell say, $c_w$ which either contains $dot_i$ and $dot_j$; \\
				     Add the dot which is not the part of Braille cell $c_w$; \\
			    } 
		    	{\textit{\{Hence, $dot_i$ and $dot_j$ are part of different Braille cells say, $c_w$ and $c_x$\}} \\
		    	   Merge both the Braille cells $c_w$ and $c_x$;
		    	}
	    	}

	}}	}
	\textit{Make those dots ($dot_i$ $\notin C$)  as individual Braille cells} 
	}}
	
		\caption{Braille cell clustering}\label{euclid}
\end{algorithm} 
The proposed clustering algorithm is effective in handling the induced skewness as illustrated in Figure \ref{fig:20}. 

\subsection{Cell Transcription}  	   

The focal point of the final step is \textit{`Cell Transcription'} which extract features from Braille cells and classify them into readable natural language characters. The previous step \textit{`Cell Recognition'}, empowered us to calculate a set of Braille cells $C$ such that each Braille cell $c_i$ is cluster of Braille dots (refer Equation \ref{e6}). Now, the task at hand is to select appropriate features which can be used for efficient classification of each Braille cell. The strategy adopted by us for feature selection exploits the traits of Braille cell to regularize the placement of dots inside the cell. This implies that there exits one-to-one relationship between physical structure of dots and natural readable characters. The most important physical parameter to derive our feature vector is \textit{`centroid'} of any Braille cell. Its importance can be gauged by its ability to enable the estimation of dot placement inside a cell using the distance between its centroid and the dot. 
\subsubsection{Centroid Estimation:} 
\begin{figure}
	\centering
	
	\includegraphics[height=0.15\textheight, width=0.8\linewidth]{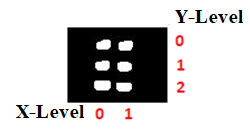}
	\caption{ X and Y level for a given Braille cell}
	\label{fig:13}       
\end{figure}

The Braille cells are organized in an array of $2 \times  3$ i.e., there are two horizontal levels (X-level) and three vertical levels (Y-level) as shown in Figure \ref{fig:13}.  If we consider Braille cell to be rectangular in shape then finding centroid of a Braille cell is a straight forward task.
\begin{figure}
\centering

\includegraphics[scale = 0.3]{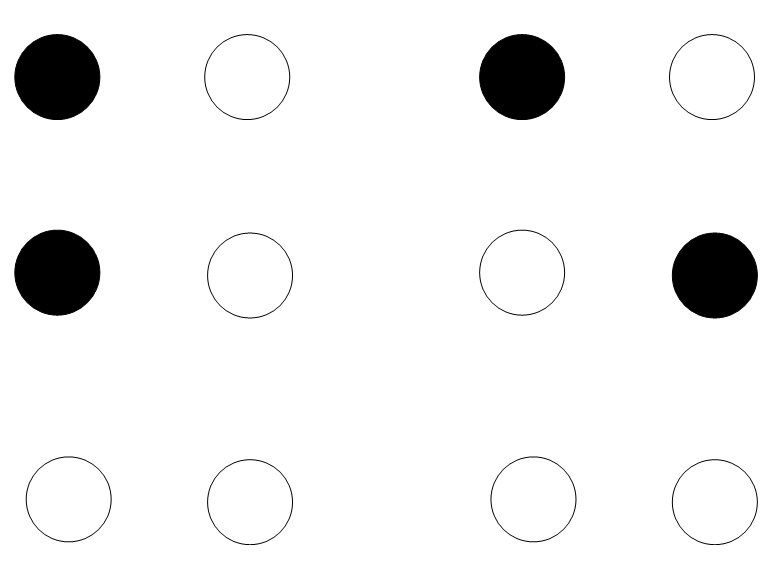}
\caption{ Two Braille cell representing word `be'.}
\label{fig:be} 
\end{figure}
\begin{figure}
	\centering
	
	\includegraphics[height=0.1\textheight, width=0.2\linewidth]{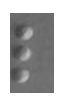}
	\caption{ A Braille cell representing word \textit{`l'}.}
	\label{fig:la} 
\end{figure}  
However, for the cases like letter `A' which contains one dot, the centroid is displaced as a centre of existing dot (illustrated in Figure \ref{fig:9}).    
\begin{figure}
	\centering
	
	\includegraphics[height=0.13\textheight, width=0.8\linewidth]{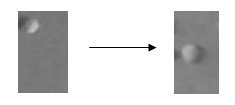}
	\caption{(a) Correct centre detected (b) Displaced centre detected }
	\label{fig:9}       
\end{figure}

Therefore to overcome the paradox of centriod shifting, we have proposed a robust centroid detection algorithm. The proposed algorithm can detect correct centroid position irrespective of the number of dots. Algorithm \ref{euclid3} is assisted in this task using two functions \textit{Correct\_X( )} and \textit{Correct\_Y( ) } which are algorithmically described in Algorithm \ref{euclid1}. The intuition behind these two functions have been discussed in details in upcoming section. The centroid detection algorithm (Algorithm \ref{euclid3}) takes a set of Braille cells $C$ (calculated using algorithm \ref{euclid}) as an input and generates a set of centroid (denoted as $CENT$).  

This algorithm also requires  two more inputs which are average inter-cell horizontal and vertical distances, respectively. These distances have been calculated using histogram analysis of  $HOR_{near}$ and $VER_{near}$. The histogram analysis of  $HOR_{near}$ and $VER_{near}$ was carried out in previous section (refer Section 2.2) to calculate maximum intra-cell distance. $HOR_{near}$ and $VER_{near}$ stores minimum inter-dot distances horizontally and vertically, respectively (calculated using Equation \ref{eq:1} and \ref{eq:2}). 

The closest dot to any given dot will belong to the same cell. But for few words like \textit{`be'} (illustrated in Figure \ref{fig:be}), the closest dots horizontally belong to different Braille cells. Presence of such words in any given document will allow second peak of the histogram to emerge. However, the first peak corresponds to the average intra-cell dot distance. The inter-cell dot distance is calculated as difference between the first peak and the second peak of the histogram. 

 $HOR_{inter}$ denotes inter-cell dot distance which are calculated from histogram of $HOR_{near}$. Similarly, the vertical inter-cell dot distance will be calculated from histogram of $VER_{near}$ and will be denoted as  $VER_{inter}$. 
\subsubsection{Centroid Correction:} 
In this section, we will describe the intuition behind two aforementioned functions \textit{Correct\_X( )} and       \textit{Correct\_Y( ) } in rectification of centroid detection of the Braille cells, if there exists any irregularities.
The function \textit{Correct\_X( )} corrects the displaced x-coordinate of centroid of the cell due to absence of dots at the two X-levels. In order to achieve this feat, we detect the level at which the dot is absent. The given cell is modified by adding a dot at remaining X level. This allows us to rectify the x-coordinate of centroid. Similar operation is performed for correction of y-coordinate of centroid using function \textit{Correct\_Y( )}. Unlike the previous scenario of x-coordinate correction, y-coordinate correction arises due to the absence of dots at any three Y-levels in any given Braille cells. Thus, the function \textit{Correct\_Y( )} adopts a similar strategy by adding dots at missing Y-levels for correct calculation of centroid. 
\begin{algorithm} [!h]
	
	\small{
		\KwIn{ \\ 1. $C$ : a set of clusters on Braille cells ; \\ 
			2. $HOR _{inter}$ , $VER_{inter}$ : Horizontal and Vertical inter-cell dot distances, respectively; \\

		}
		\KwOut{ $CENT = \{cent_1\dots cent_w\}$ : Set of centroid of Braille cells, $C$ }
		
		\textbf{Initialization:} \\
		
		1. $CENT = NULL; $ \\
		2. $Sample =NULL;$  \Comment{Reference cell} \\
		\For{each cell $c_i \in C$}{
			\If{ there exists dots in all three Y levels and two X levels (refer Figure \ref{fig:13}) }{
				$cent_i = $ centroid of all $dots_j \in c_i$; \\
				$Sample = c_{i}$; \\
				\Comment{Store any one cell in $Sample$ for later correction.} 
			}

			\If{ there exists dots in all three Y levels only}{
				
				\textit{\{Since, there are dots present at all the three Y-levels, y-coordinate of centroid calculated is correct and only x-coordinate needs to be corrected.\}} \\
				$cent_i^y = $ y-coordinates of centroid of all $dots_j \in c_i$; \\
				$cent_i^x = Correct\_X(c_i,Sample,$ $ HOR_{inter})$;
			}
			\If{there exists dots in all two X levels only}{
				
				\textit{\{There are dots present at all two X-levels, x-coordinate of centroid calculated is correct and only y-coordinate needs to be corrected.\}} \\
				$cent_i^x = $ x-coordinates of centroid of all $dots_j \in c_i$; \\
				$cent_i^y = Correct\_Y(c_i, Sample,$ $VER_{inter})$;
			} 
			\If{there exists dots \textbf{not} in two X levels and Y levels}{
				\textit{\{ Insufficient dots present at all the levels of Y, both x and y coordinates needs to be corrected.\}} \\
				$cent_i^x =$ $Correct\_X(c_i ,Sample$, $HOR_{inter})$; \\
				$cent_i^y = $ $ Correct\_Y(c_i,Sample$, $VER_{inter})$; 
			}
	}    } 
	\caption{Centroid detection}\label{euclid3}
	
\end{algorithm}
\begin{algorithm} [!h]
	\SetKwFunction{FMain}{Correct\_ X ( )}
	\SetKwProg{Fn}{Function}{:}{}
	\Fn{\FMain}{
		
		\textit{\{Calculate horizontal distances between any dot, $d$,  in $c_i$ and both dots ($dot_1,dot_2$) present at two X-levels in $Sample$\}} \\
		$h_1 = H_D(dot_1,d)$ and $h_2= H_D(dot_2,d)$; \\ \Comment{Using equation \ref{e3}}; \\
		\eIf{$h_1~mod~HOR_{inter} < h_2~mod~HOR_{inter} $}{
			\textit{\{Dot $d$ lies at the $X$ level of $dot_1$\}} \\
			Add another dot $d'$ to cell $c_i$ at the $X$ level of $dot_2$ with same $Y level$ as dot $d$;    
		}
		{
			\textit{\{Dot $d$ lies at the $X$ level of $dot_2$\}} \\
			Add another dot $d'$ to cell $c_i$ at the $X$ level of $dot_1$ with same $Y level$ as dot $d$;
			
		}
		Q = x-coordinates of centroid of modified cell $c_i$; 
		\KwRet Q 
	}
	
	\SetKwFunction{FMain}{Correct\_ Y ( )}
	\SetKwProg{Fn}{Function}{:}{}
	\Fn{\FMain}{
		
		\textit{\{Calculate vertical distances between any dot($d$) in $c_i$ and all three dots ($dot_1,dot_2,dot_3$) present at two level in $Sample$\}} \\
		$v_1 = V_D(dot_1,d)$, $v_2= V_D(dot_2,d)$ and $v_3 = V_D(dot_3,d)$; \\ 
		\Comment{Using Equation \ref{e4}}; \\
		\eIf{$v_1~mod~VER_{inter}$ is min($v_1~ mod~VER_{inter}$, $ v_2~mod~VER_{inter}$,$v_3~mod~VER_{inter}) $)}{
			\textit{\{Dot $d$ lies at the $Y$ level of $dot_1$\}} \\
			Add dots remaining Y-level of cell $c_i$;    
		}
		{
			\eIf{$v_2~mod~VER_{inter}$ is min($v_1~mod~$$VER_{inter}$, $v_2~mod~VER_{inter}$,$v_3~mod~VER_{inter}$) )}{
				\textit{\{Dot $d$ lies at the $Y$ level of $dot_2$\}} \\
				Add dots remaining Y-level of cell $c_i$;  }{
				\textit{\{Dot $d$ lies at the $Y$ level of $dot_1$\}} \\
				Add dots remaining Y-level of cell $c_i$;    }  
			
		}
		Q = y-coordinates of centroid of modified cell $c_i$; 
		\KwRet Q 
	}
	\caption{Function Correct\_X( ) and \qquad Correct\_Y( )}\label{euclid1}
\end{algorithm} 
\subsubsection{Feature Extraction}
After detecting centroid (using Algorithm \ref{euclid3}), we are equipped to extract features to complete the last step,  `Cell Translation'. As per our in-depth discussion in earlier section (Section 2.3), the identified features are unique characteristics of each Braille cell. The feature vector ($FV_{c_i}$) representing a Braille cell, $c_i$ can be mathematically formulated as:
\begin{equation}
\qquad \qquad \qquad FV_{c_i} = \{~N_{c_i}, Dot\_Code_{c_i}~\}
\end{equation}
where,
\begin{itemize}
	\item $N_{c_i}$ is the number of the raised dots in a Braille cell. For instance, in case of  letter \textit{`l'}  (illustrated in Figure \ref{fig:la}) $N_{c_i}$ will be three.
	\item $Dot\_Code_{c_i}$ denotes a Braille cell as a set of tuples, where each tuple represents the occurrence position of dots in cell, $c_i$. 
\end{itemize}
The occurrence positions of dots are indexed as X-level and Y-level. $Dot\_Code_{c_{i}}$ for any cell $c_i$ can be computed as:
\begin{equation}
\qquad  Dot\_Code_{c_i} = ~\{\langle X_{dot_j},Y_{dot_j} \rangle |~ \forall dot_j \in c_i \}
\label{e9}
\end{equation}       
For example, if we look at the same letter \textit{`l'} again as  illustrated in Figure \ref{fig:la}. There are three dots $d_1, d_2,  d_3$  occurring at the first X-level, therefore as per equation \ref{e9}, $Dot\_Code_{l}$  is equivalent to $\{<0,0,>,<0,1>,<0,2>\}$. The extraction of feature vector $FV_{c_i}$ for a Braille cell $c_i$ is performed using Algorithm \ref{algo1}. Algorithm \ref{algo1} uses `position of centroid' as a reference point to determine the X-level and Y-level of dots for a cell $c_i$.  

\begin{algorithm}[!h]
	\small{
		\KwIn{\\1. $c_i$ : a given Braille cell; \\
			2. $cent_{c_i}$ : centroid of Braille cell $c_i$; 
		}
	\KwOut{$FV_{c_i}$ : feature vector of Braille cell $c_{i}$
	}
    \Begin{
    		\textbf{Initialization:} \\
    		 1. $N_{c_i} = |c_i|; $ 
    		\Comment{$N_{c_i}$ is number of dots in cell $c_i$.} \\
    		2. $Dot\_Code_{c_i} = NULL; $ 
    		\Comment{$Dot\_Code_{c_i}	$ is indexing of dots in cell $c_i$. \\
    		\For{ each $dot_j \in c_i$ }{
    		 Calculate: \\
    		     \qquad $v~=~dot_j^y - cent_{c_i}^y$; \\
    		     \qquad $h~=~dot_j^x - cent_{c_i}^x$; 
    		 \Comment{Vertical and Horizontal distances between dot $dot_j$ and centroid $cent_{c_i}$.} \\
    		 \eIf{ v ~ = 0}{
    		 	$Y_{dot_j}$ = 1 ; 
    		 	\Comment{$dot_j$ lies at the same Y-level as centroid.}
    		 }{
    	    \eIf{ v is positive}{
    	    	$Y_{dot_j}$ = 2 ; 
    	    	\Comment{$dot_j$ lies below centroid.}
    	    }
        	{
        	$Y_{dot_j}$ = 0 ; 
        	\Comment{$dot_j$ lies above centroid.}
        	} }
        \eIf{ h is positive}{
        	$X_{dot_j}$ = 1 ; 
        	\Comment{$dot_j$ lies to the right of centroid.}
        }
        {
        	$X_{dot_j}$ = 0 ; 
        	\Comment{$dot_j$ lies to left of centroid.}
        }
         $Dot\_Code_{c_i} = $ $Dot\_Code_{c_i}~\cup~ \langle X_{dot_j},Y_{dot_j} \rangle $ ;
        	}
    		$FV_{c_i} = \{ N_{c_i}, Dot\_Code_{c_i} \}$;
    		\Comment{Concatinating $N_{c_i}$ and  $Dot\_Code_{c_i}$ to form feature vector $FV_{c_i}$.}	
    	}    
	}
}
\caption{Feature extraction} 
\label{algo1}
\end{algorithm}

After extraction of feature vector, an ensemble learning method \textit{`random forest'} \cite{ho1998random} is used for classification of the Braille cells. Random forest  usually searches a space of hypothesis to find the best hypothesis by constructing multiple individual decision trees at the time of training. This typical behaviour makes it an ideal classification technique to avoid the problem of overfitting \cite{rocha2017naive}. In case of Braille recognition, there is absence of large training datasets. Therefore, the problem of overfitting is quite persistent in OBR  algorithms. Consequently, using random forest can overcome this glitch due to its generalization capabilities.  

We have applied the technique of bootstrap aggregation \cite{breiman2001random} to individual decision trees to further increase the accuracy. This allows random forest to outperform many other existing classification techniques reducing the variance by averaging over learners  and randomising the stages to decrease the correlation between distinctive learners in the ensemble. Using random forest, we output class labels to the input feature vectors. The next section will present our experimental setup and evaluation of our proposed algorithm on the collected dataset.       
\section{Experiments and Results}
\begin{figure}
	\centering
	\includegraphics[height= 0.2\textheight , width=\linewidth]{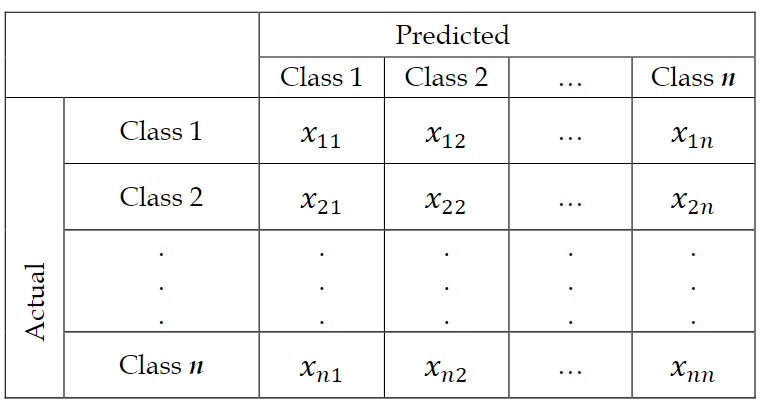} \\
	\caption{Confusion matrix for n classes}
	\label{fig:7}
\end{figure}

Performance evaluation of our OBR algorithm has been carried out through a comprehensive set of experiments on the proposed dataset captured using a portable system where the camera of smartphones is used to capture Braille documents. The Confusion Matrix, Sensitivity, Specificity and Accuracy are the principal metrics based on which the evaluation of our algorithm has been performed. The analysis has been conducted for two levels, dot detection and Braille recognition. At first, we will define the primary components of all the metrics. 
\begin{enumerate}
	\item True Positive (TP) : An outcome is classified as TP if a raised Braille dot is correctly identified. For Braille recognition, TP of a given character is the number of cells correctly labelled. From Figure \ref{fig:7}, TP for a given character $k$  can be calculated as:  
	\begin{equation}
	\qquad \qquad \qquad TP_k =  x_{k,k}
	\end{equation}
	\item True Negative (TN) : An outcome is classified as TN if a flat Braille dot is correctly identified. For Braille recognition, TN for a given character is the number of cells excluding the cells incorrectly or correctly labelled as the given character. From Figure \ref{fig:7}, TN for a given character $k$  can be calculated as:  
	\begin{equation}
	\qquad \qquad \qquad TN_k = \sum_{i,j= 1|i,j \ne k} ^{n} x_{i,j}
	\end{equation}
	\item  False Positive (FP) : An outcome is classified as FP if a depressed Braille dots is identified as a flat dot.
	For Braille recognition, FP for a given character is the number of cells incorrectly labelled as the given character. From Figure \ref{fig:7}, FP for a given character $k$  can be calculated as:  
	\begin{equation}
	\qquad \qquad \qquad FP_k = \sum_{i= 1 | i \ne k}^{n} x_{i,k}
	\end{equation}
	\item False Negative (FN) :  An outcome is classified as FN if a flat Braille dots is identified as a depressed dot. For Braille recognition, FN for any given character is the number of cells incorrectly labelled as some other character. From Figure \ref{fig:7}, FN for a given character $k$  can be calculated as:  
	\begin{equation}
	\qquad \qquad \qquad FN_k = \sum_{j=1|j\ne k}^{n} x_{k,j}
	\end{equation}
\end{enumerate}

The confusion matrix is a tabular representation of the performance of classification model. The confusion matrix for Dot detection can be mathematically stated in Table \ref{tab:1}. The confusion matrix for Braille cell recognition can be formulated as shown in Figure \ref{fig:7}.
The remaining metrics are defined as:
\begin{table}
	\caption{Dot detection confusion matrix}
	\centering
	\label{tab:1}       
	\begin{tabular}{c c c}
	\hline\noalign{\smallskip}
	Dot & Detected &  Not Detected  \\
	\noalign{\smallskip}\hline\noalign{\smallskip}
	Protrusion & TP & FP \\
	Flat & FN & TN \\
\noalign{\smallskip}\hline
\end{tabular}
\end{table}

\begin{equation}
\qquad \qquad \qquad Specificity = \frac{TN}{TN + FP}
\label{r1}
\end{equation}
\begin{equation}
\qquad \qquad \qquad Sensitivity = \frac{TP}{TP + FN}
\label{r2}
\end{equation}
\begin{equation}
\qquad \qquad Accuracy = \frac{TP + TN}{TP + TN + FN + FP}
\label{r3}
\end{equation}
Before presenting the detailed results, we now discuss our proposed dataset.
\subsection{Dataset}
The main motivation of this work is to render assistance to blind students by bridging the written communication gap between them and non-Braille users. Thus, a deliberate effort has been made to translate their personal academic documents  like assignments and exam papers. Hence, our proposed datasets contains 54 English Grade 1 single-sided Braille documents. The dataset compromises of academic texts with upper and lower case letters, numerals and other symbols like punctuation marks. This content has been assembled from online websites \cite{wiki:xxx}. The annotation and labelling of dataset was performed by two literate users. One of the Braille user was child and other user was sighted. This permits incorporation of perception of both the type of Braille users.
  
The portability of algorithm is ensured by capturing all the Braille pages using a camera of smartphone. The detailed characteristics of this dataset has been tabulated in Table \ref{tab:2}.
\begin{table}[t]
	\caption{Characteristics of a system}
	\centering
	\label{tab:2}       
	\begin{tabular}{c c}
		\hline\noalign{\smallskip}
		\textbf{Features} &\textbf{Numbers} 
		\\
		\noalign{\smallskip}\hline\noalign{\smallskip}
		Number of Braille Pages & 54\\
		Number of Braille Dots  & 39232 \\
		Number of Braille Cells & 13188 \\
		Digital format & RGB \\
		Resolution & 96 dpi (horizontal and vertical) \\
		Image size & 250 Kbytes \\
		Image format & JPEG (.jpg) \\
		Braille type & Single sided \\
		Document size & 26.5 cm. (horizontal) x 32 cm. (vertical) \\
		Smartphone & Samsung galaxy grand duos \\
		Camera resolution & 8 megapixel \\
		\\
		\noalign{\smallskip}\hline
	\end{tabular}
\end{table}  

\subsection{Results and Analysis}
The performance evaluation of our OBR algorithm has been carried out by measuring all the principal metrics  on the dataset discussed above. We will now evaluate our experimental observation based on these metrics at two fronts i.e. dot detection and Braille recognition in the subsequent sections.
\subsubsection{Dot Detection}
The proposed algorithm is a three step process which computes many parameters to cluster Braille cells and classify them into natural language characters. However, all the parameters are dependent on the ability to detect Braille dots. Table \ref{tab:3} shows the obtained confusion matrix for dot detection. From this confusion matrix, we derive all the primary components (ie. TN, TP, FN, FP) which serve as a building block to estimate other metrics. Now, using equations \ref{r1}, \ref{r2}, \ref{r3}, we have calculated Specificity, Sensitivity and Accuracy and depicted it in Table \ref{tab:4}.    
\begin{table}[t]
	\caption{Confusion matrix for  dot detection}
	\centering
	\label{tab:3}   
\begin{tabular}{c  c c  | c}
	\hline\noalign{\smallskip}
	\textbf{Dot}&\textbf{Protrusion} &\textbf{Flat}&\textbf{Total} 
	\\
	\noalign{\smallskip}\hline\noalign{\smallskip}
	Protrusion & 39226 & 6 & 39232 \\
	Flat & 8 &39861  & 39869 \\
	\noalign{\smallskip}\hline\noalign{\smallskip}
	Total & 39234 &39867 &79101 \\
	\\
	\noalign{\smallskip}\hline
\end{tabular}
\end{table}
\begin{table}[t]
	\caption{Principal metrics for  dot detection}
	\centering
	\label{tab:4}   
	\begin{tabular}{c  c}
	\hline\noalign{\smallskip}
	\textbf{Specificity}& 0. 9998 \\
	\noalign{\smallskip}\hline\noalign{\smallskip}
	\textbf{Sensitivity} &  0.9998 \\
	\noalign{\smallskip}\hline\noalign{\smallskip}
	\textbf{Accuracy} & 0. 998  \\

\noalign{\smallskip}\hline
\end{tabular}
\end{table}
It is evident from Table \ref{tab:4} that high values of each metrics specifically 99.8 \% accuracy validate the proposed dot detection mechanism. 

This experimental result substantiates techniques deployed in first two steps of our OBR algorithm.  
If we further throw light, a series of pre-processing techniques used successfully eradicates the undesired effects of artifices like skewness and noise. Another major point to consider is use of standard circular shape of Braille dots for their identification is an apt choice. The closer look at the metric accuracy elucidates that detection of dots using Hough transform is appropriate due to its ability to identify miniature sized circular objects.
\subsubsection{Braille Recognition}
\begin{figure}
	\centering
	\subfloat[]{
		\includegraphics[width=0.5\linewidth]{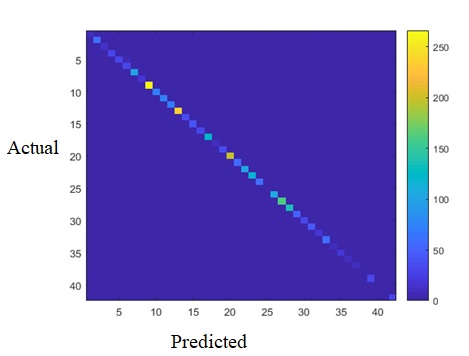}} 
	\subfloat[]{
		\includegraphics[width=0.5\linewidth]{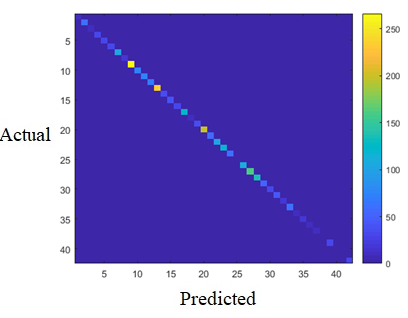}}\\
	\subfloat[]{
		\includegraphics[width=0.5\linewidth]{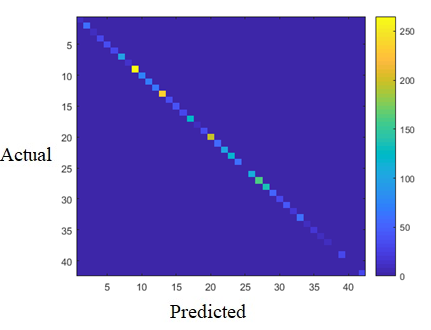}} 
	\subfloat[]{
		\includegraphics[width=0.5\linewidth]{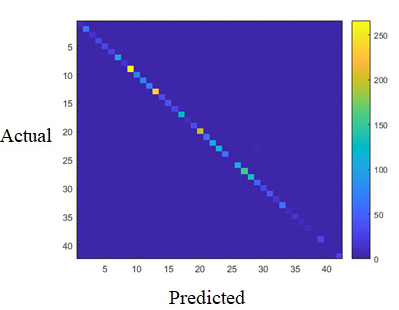}}\\
	\subfloat[]{
		\includegraphics[width=0.5\linewidth]{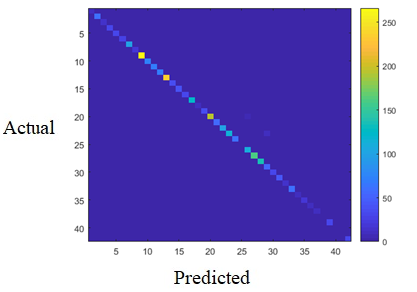}}
	\caption{Confusion matrix plots for all the 5 folds}
	\label{fig21}
\end{figure}

\renewcommand{\arraystretch}{.7}
\renewcommand{\tabcolsep}{11pt}
\begin{table*}[t]
	\caption{Comparison with other techniques}
	\centering
	\label{tab:7}       
	\begin{tabular}{|c c c| }
		\hline\noalign{\smallskip}
		\textbf{Existing Work by}&\textbf{Achieved Accuracy} &\textbf{Employed Methodology}
		\\
		\noalign{\smallskip}\hline\noalign{\smallskip}
		Li T et al. \cite{li2014deep} &92\%&  Autoencoder based feature extraction from pre-segmented Braille cells. \\
		Antonacopoulos et al. \citep{antonacopoulos2004robust}  & 94.9\%-99\% & Thresholding based dot localization and grid overlaying  \\
		Yousefi et al. \citep{yousefi2012robust} & 96.8 \% & maximum-likelihood based parameter estimation of Braille cells \\
		Namba et al. \citep{namba2006cellular} & 87.9\% & Cellular Neural Network based Braille cell recognition\\
		Z. Tai et al. \citep{tai2010braille} & 82.7\% & Belief Propogation based Braille recognition \\
		Khanam et al. & \textbf{98.71\% }& Hough transform based dot detection\\  
		\noalign{\smallskip}\hline
		
	\end{tabular}
\end{table*}
In previous section, we scrutinized the first two steps of our algorithm. Now, we will access the last step which will in turn cover the performance of our  algorithm. We measure all the metrics using 5-fold cross validation. The proposed dataset is partitioned into  two parts with 80 percent used as training data , while 20 percent data was used for testing. The principal metrics like Accuracy, Sensitivity and  Specificity are calculated after each fold and listed in Table \ref{tab:6}. Alongside, we plot  confusion matrix for each fold which is illustrated using Figure \ref{fig21}. 
\renewcommand{\arraystretch}{.9}
\renewcommand{\tabcolsep}{5pt}
\begin{table}[h]
	\caption{Results of 5-fold cross validation}
	\centering
	\label{tab:6}       
	\begin{tabular}{c c c c c}
		\hline\noalign{\smallskip}
		\textbf{Fold}&\textbf{Accuracy} &\textbf{Error}&\textbf{Sensitivity}& \textbf{Specificity} 
		\\
		\noalign{\smallskip}\hline\noalign{\smallskip}
		1  & 0.9833 & 0.0117 & 1 & 1\\
		2  & 0.9870 & 0.0130 & 1 & 1\\
		3  & 0.9903 & 0.0097 & 1 & 1\\
		4  & 0.9906 & 0.0094 & 1 & 1\\
		5  & 0.9841 & 0.0159 & 1 & 0.9996\\
			\noalign{\smallskip}\cline{1-5} \\
		 \textbf{Overall: }& 0.9871 &0.0129 & & \\
	\hline\noalign{\smallskip}

	\end{tabular}
\end{table}

A closer examination at the experimental value obtained for all the principal metrics can allow us to derive the following conclusions. The high TP value for most of the classes (natural language characters) across each fold indicates that most Braille cells are correctly characterized. From the low FN and FP values, one can infer that there exists rare case of Braille cell misclassification.

The previous subsection has affirmed our dot detection mechanism. The result presented here is an attestation of steps mainly: Braille cell clustering and classification. Algorithm \ref{euclid} efficiently clusters detected dots as Braille cells which prompts accurate deduction of parameters like inter-cell and intra-cell dot distances. The accurate estimation contributes to selection of robust features and classification.

\subsubsection{Comparison with Existing Works}

In Table \ref{tab:7}, we have made a comparision with other state-of-the-art techniques using the accuracy metric. All the existing methods have evaluated their technique using accuracy metric as stated in Equation \ref{r3}. The accuracy metric is a lucrative choice for evaluation of any classification technique with respect to its counterpart. Thus in the given scenario of Braille recognition, it is closely realistic in nature as a comparison metric.   

It is evident from the table that our approach comprehensively outperforms the existing techniques. This may be attributed to strong dot detection mechanism and robot feature selection. It is worth observering that each of the algorithms are implemented on different dataset, therefore this comparison should be taken by a pinch of salt. However, one can easily derive from the trends of high accuracy numbers that this algorithm performs at-par.  Apart from that, robustness of the algorithm to cluster dots into Braille cell and classify them can also be perceived from achieved accuracy of 98.71\%with proposed dataset of 54 Braille documents.
 \section{Conclusion and Future Work}
 In this work, we proposed a new technique for Optical Braille Recognition which facilitates the conversion of personal academic Braille scripts into readable natural language. The proposed technique is carried out as a three-step process: Digitization, Cell Recognition and Cell Transcription. The first step, \textit{`Digitization'} comprises of image acquisition of Braille scripts followed by application of series of pre-processing techniques. The next step, \textit{`Cell Recognition'}, detects the Braille dots using Hough transform. The last step, \textit{`Cell Transcription'}, extracts the robust features from Braille cells and classifies them into natural language characters. The comprehensive analysis has revealed the efficacy of proposed technique.
 
The main motivation of this work was to assist blind students in translation of their personal academic documents from Braille to natural language. Thus, we have proposed a dataset containing English Grade 1 single-sided Braille scripts. The promising results achieved using our proposed dataset is a strong incentive to implement similar mechanism tailored for double-sided Braille scripts. Hough transform based strong dot detection mechanism achieved 99.8\% accuracy on the proposed dataset has affirmed the hypothesis of using circular shape as a prime criteria to detect Braille dots. Thus, the proposed technique can easily be extended to double-sided Braille documents in future. It is also worth pursuing the application of this approach to Braille scripts of other languages and domains. 

\section*{ACKNOWLEDGEMENT}
We acknowledge the support of Ahmadi School for the Visually Challenged, Aligarh Muslim University in collection of Braille Scripts. We recognise
Dr. Sangeet Saha for his suggestion during manuscript writing phase.


\bibliographystyle{spbasic}      
\bibliography{exam1} 


%
%

\end{document}